\documentclass[sigconf]{acmart}
\usepackage{multirow}
\usepackage{booktabs}
\usepackage{graphicx}
\newcommand{\std}[1]{{\scriptsize $\pm$#1}}

\AtBeginDocument{%
  }

\copyrightyear{2026}
\acmYear{2026}
\setcopyright{cc}
\setcctype{by}
\acmConference[CIKM '26] {Proceedings of the 35th ACM International Conference on Information and Knowledge Management}{November 7--11, 2026}{Rome, Italy.}
\acmBooktitle{Proceedings of the 35th ACM International Conference on Information and Knowledge Management (CIKM '26), November 7--11, 2026, Rome, Italy}
\acmISBN{979-8-4007-2539-5/2026/11}
\acmDOI{10.1145/3799682.3840948}  

\begin{document}

\title{Multi-View Molecular Representation Learning with Hierarchical Graphs and Contextualized Fingerprints}

\author{Gwang-Hyeon Yun}
\authornote{These authors are contributed equally to this work.}
\orcid{0009-0008-4180-5776}
\affiliation{
  \institution{Yonsei University}
  \city{Wonju-si}
  \country{Republic of Korea}
}
\email{ghyun0130@yonsei.ac.kr}

\author{Jong-Hoon Park}
\authornotemark[1]
\orcid{0000-0002-8774-5640}
\affiliation{%
  \institution{Yonsei University}
  \city{Wonju-si}
  \country{Republic of Korea}
}
\email{jonghoon\_park@yonsei.ac.kr}

\author{Bing Hu}
\orcid{0009-0006-2913-6659}
\affiliation{%
  \institution{University of Waterloo}
  \city{Waterloo}
  \country{Canada}
}
\email{b25hu@uwaterloo.ca}

\author{Helen Chen}
\orcid{0000-0002-9401-7895}
\affiliation{%
  \institution{University of Waterloo}
  \city{Waterloo}
  \country{Canada}
}
\email{helen.chen@uwaterloo.ca}

\author{Anita Layton}
\orcid{0000-0002-1753-4063}
\affiliation{%
 \institution{University of Waterloo}
 \city{Waterloo}
 \country{Canada}
}
\email{anita.layton@uwaterloo.ca}

\author{Young-Rae Cho}
\authornote{Corresponding author.}
\orcid{0000-0002-4645-2542}
\affiliation{%
  \institution{Yonsei University - Mirae Campus}
  \city{Wonju-si}
  \country{Republic of Korea}
}
\email{youngcho@yonsei.ac.kr}

\renewcommand{\shortauthors}{Gwang-Hyeon Yun et al.}

\begin{abstract}
  Molecular property prediction requires representations that generalize from limited labeled data to structurally novel compounds. Existing molecular pretraining methods often rely on a single view: graph-based approaches model atom-bond topology but provide limited fragment-level supervision, whereas fingerprint descriptors encode chemical patterns but are typically used as fixed auxiliary features. We propose \textbf{HiFi-Mol}, a multi-view framework that separately pretrains a \textbf{hi}erarchical graph encoder and a contextualized \textbf{fi}ngerprint encoder before downstream integration. The graph branch uses fragment-aware masking with multi-resolution supervision to capture substructure-aware representations, while the fingerprint branch tokenizes active entries from seven fingerprint families and applies masked language modeling to learn contextualized embeddings. During fine-tuning, HiFi-Mol combines projected multi-resolution graph features with fingerprint embeddings for downstream prediction. Evaluated on MoleculeNet benchmarks under the scaffold split, HiFi-Mol achieves a 2.77\% improvement in average ROC-AUC over the best baseline across eight classification tasks while maintaining competitive performance on three regression tasks. Further analyses reveal that fragment-aware masking improves graph representation quality, and classification results demonstrate dataset-dependent strengths of the individual graph and fingerprint variants, confirming that the two views provide complementary predictive signals.
\end{abstract}




%\received{24 May 2026}
%\received[revised]{23 August 2026}
%\received[accepted]{5 June 2026}

\maketitle

\section{Introduction}
Molecular property prediction~\citep{stark20223d, chu2026memol} is a central problem in early-stage drug discovery, where computational models are used to prioritize promising compounds and filter toxic or inactive candidates before costly experimental validation~\citep{stokes2020deep, liu2025molecular, qiao2025self}. Despite significant progress, two key challenges remain. First, labeled molecular data are often scarce, forcing models to learn from limited supervision~\citep{li2025contextual}. Second, such models must generalize to structurally novel molecules beyond the training distribution, since newly evaluated compounds are often absent from the training data~\citep{antoniuk2026boom, liu2025subgraph}. These challenges place strong demands on molecular representation learning: useful representations should encode not only atom-level topology but also higher-order chemical structures and functional patterns that govern molecular activity.

Graph neural networks (GNNs) have become a dominant approach for molecular representation learning because molecules naturally form atom-bond graphs~\citep{gilmer2017neural, hu2019strategies, rong2020self}. 
However, atom-level graph representations mainly capture local relational topology through message passing over atoms and bonds, making it difficult to explicitly model chemically meaningful substructures such as functional groups, ring systems, and pharmacophoric scaffolds~\citep{jiang2023pharmacophoric}.
While recent methods such as MGSSL~\citep{zhang2021motif} and HiMol~\citep{zang2023hierarchical} incorporate fragment-level nodes, they either decouple fragment identities from their constituent atom-bond patterns or lack multi-resolution supervision spanning atom, fragment, global, and geometric signals.

Molecular fingerprints provide another chemically informed view of molecular structure~\citep{yang2019analyzing, david2020molecular, atz2021geometric}. 
They encode predefined descriptors such as circular substructures, pharmacophoric patterns, and topological relations. 
Although many fingerprint descriptors are derived from molecular topology, their discrete and predefined encoding scheme can capture structural patterns that are complementary to learned graph representations~\citep{deng2023systematic}.
However, existing molecular pretraining methods rarely treat these two views as learnable, multi-view components within a unified framework: fingerprints are typically incorporated as static auxiliary vectors and shallowly concatenated with graph embeddings, limiting their ability to model contextual descriptor relationships. 

Motivated by these observations, we propose HiFi-Mol, a multi-view molecular pretraining framework composed of a \textbf{Hi}erarchical graph encoder and a contextualized \textbf{Fi}ngerprint encoder. 
Rather than enforcing a shared pretraining space, HiFi-Mol learns graph and fingerprint views with view-specific objectives and integrates them during downstream adaptation. 
On the graph side, fragment-aware masking corrupts chemically coherent fragments together with their constituent atoms and bonds, while multi-resolution supervision over atom, bond, fragment, global, and geometric signals guides the learning of hierarchical representations with fragment-level chemical semantics. 
In parallel, the fingerprint encoder represents active entries from multiple fingerprint families as descriptor tokens, which are processed together with accompanying SMILES tokens for joint masked language modeling~\citep{devlin2019bert}. This objective captures contextual dependencies among fingerprint descriptors while leveraging SMILES-derived molecular context, yielding fingerprint representations beyond static descriptor vectors.
Finally, projected graph and fingerprint representations are concatenated and passed to a task-specific predictor for downstream property prediction.

Experiments on MoleculeNet~\citep{wu2018moleculenet} benchmarks show strong classification performance and competitive regression results. 
Single-view comparisons reveal dataset-dependent strengths of graph and fingerprint representations, suggesting that the two views provide complementary predictive signals. 

The main contributions are summarized as follows:
\begin{itemize}
    \item HiFi-Mol formulates graph topology and fingerprint descriptors as complementary molecular views with distinct inductive biases, and pretrains each encoder with view-specific objectives before downstream integration.
    \item Fragment-aware masking and multi-resolution supervision jointly improve hierarchical graph representations by encouraging coherent substructure recovery across molecular resolutions. 
    \item Masked language modeling over joint SMILES--fingerprint token sequences learns contextualized descriptor representations that capture co-occurrence patterns beyond static fingerprint vectors.
\end{itemize}

\section{Related Work}
\textbf{Molecular Graph Pretraining.} Self-supervised molecular pretraining has been widely studied to alleviate label scarcity in molecular property prediction. Existing graph-based methods can be broadly categorized into generative objectives, such as AttrMask and ContextPred~\citep{hu2019strategies}, which reconstruct masked node, edge, or contextual attributes, and contrastive objectives, such as GraphCL~\citep{you2020graph}, JOAOv2~\citep{you2021graph}, and MolCLR~\citep{wang2022molecular}, which maximize agreement between augmented molecular views. Hybrid methods combine generative and contrastive pretraining signals to capture richer molecular semantics. GraphMVP~\citep{liu2021pre} and MoleculeSDE~\citep{liu2023group} incorporate geometric information by aligning 2D graph representations with 3D molecular signals, whereas Mole-BERT~\citep{xia2023mole} combines masked atom prediction with contrastive learning over multiple molecular views. While effective, these approaches typically rely on atom-level or graph-level objectives, leaving explicit supervision of chemically meaningful substructures such as functional groups, ring systems, and pharmacophores relatively underexplored.

\textbf{Fragment-level Graph Pretraining.} Recent studies have incorporated fragment- or motif-level information into molecular graph learning. MGSSL~\citep{zhang2021motif} performs motif-based generative pretraining by predicting motif labels and topological structure, while HiMol~\citep{zang2023hierarchical} constructs hierarchical molecular graphs with virtual motif and graph-level nodes. GraphFP~\citep{luong2023fragment} further aligns molecular and fragment representations through fragment-based contrastive learning. These methods demonstrate the value of modeling higher-order substructures, yet existing objectives often decouple fragment identities from the atom-bond patterns that define them.

\textbf{Fingerprint Representation Learning and Multi-View Learning.} Molecular fingerprints represent molecules as binary or count vectors encoding predefined substructural patterns, providing a complementary view of chemical structure \citep{duvenaud2015convolutional, rogers2010extended}. Prior methods such as FP-GNN~\citep{cai2022fp} and DGCL~\citep{jiang2024dgcl} suggest that combining fingerprint features with graph representations can improve downstream molecular property prediction, but fingerprints are typically used as static auxiliary descriptors and fused shallowly with graph embeddings. Recent fingerprint language modeling approaches such as DELBERT~\citep{seyed2026delbert} show that sparse fingerprint vectors can be tokenized into discrete sequences and pretrained with masked language modeling, producing contextualized descriptor representations beyond fixed fingerprint vectors. 

\begin{figure}[t]
    \centering
    \includegraphics[width=\columnwidth]{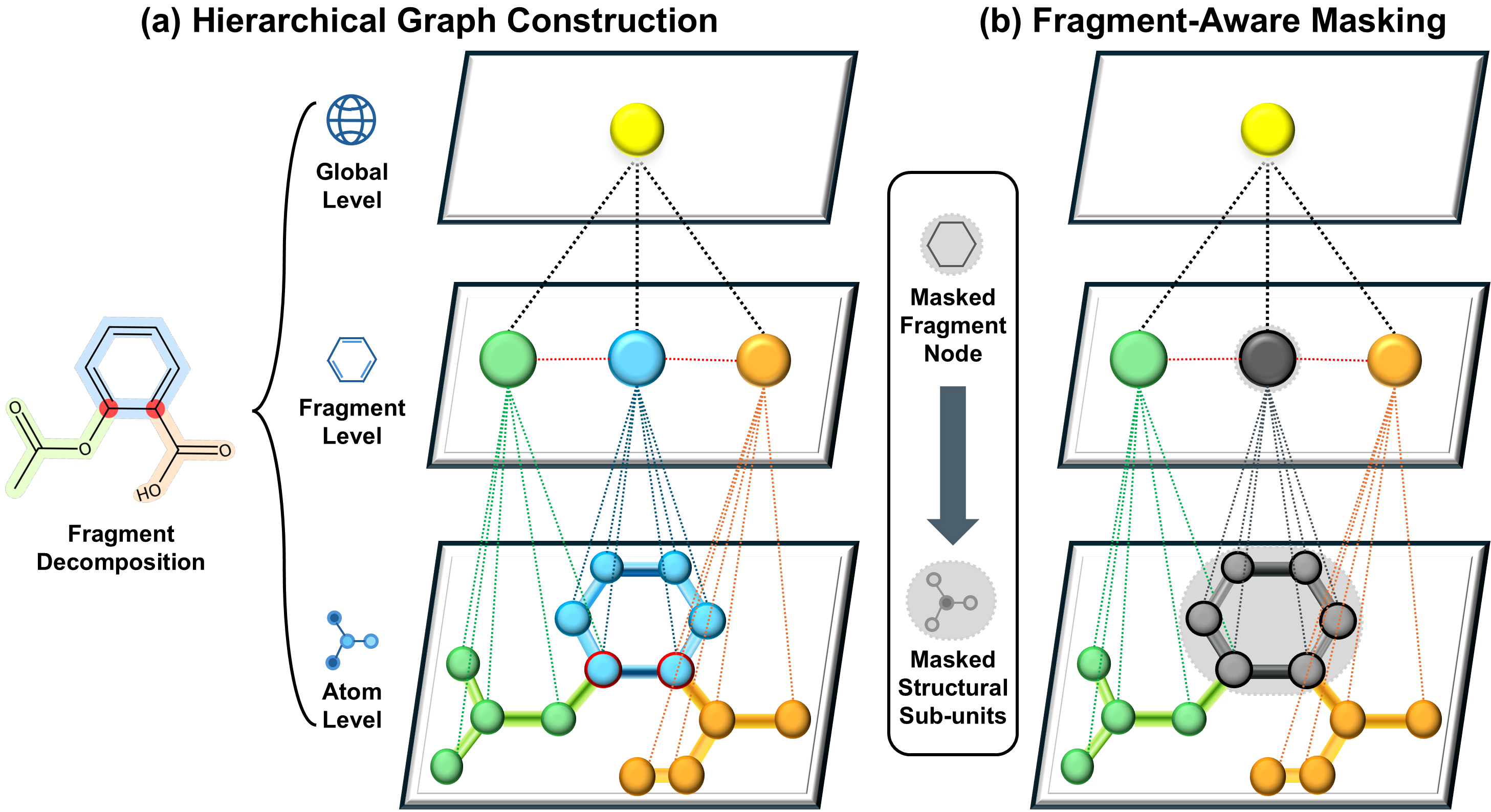}
    \caption{
        Illustration of the hierarchical molecular graph and 
        fragment-aware masking (FAM). (a) A molecule is 
        decomposed into fragment-level units using BRICS and 
        SSSR, forming a hierarchical graph with atom, fragment, 
        and global nodes. (b) FAM samples fragment nodes and 
        propagates the mask to their constituent atoms and 
        incident bonds, encouraging reconstruction of coherent 
        substructures.
    }
    \label{fig:fam}
\end{figure}

\section{Preliminaries}
\label{sec:preliminaries}
We formally introduce the two molecular views underlying HiFi-Mol and formulate the multi-view property prediction problem.

\textbf{Definition 1 (Hierarchical Molecular Graph).} 
Given a molecule $M$, its hierarchical molecular graph is defined as 
$\mathcal{G} = (\mathcal{V}, \mathcal{E})$, where
\begin{equation}
    \mathcal{V} = \mathcal{V}_{\mathrm{atom}} \cup 
    \mathcal{V}_{\mathrm{frag}} \cup \{v_\mathrm{global}\},
\end{equation}
\begin{equation}
    \mathcal{E} = \mathcal{E}_{\mathrm{bond}} \cup 
    \mathcal{E}_{\mathrm{hier}}.
\end{equation}
$\mathcal{V}_{\mathrm{atom}}$ consists of atom nodes, 
$\mathcal{V}_{\mathrm{frag}}$ consists of fragment nodes 
corresponding to chemically coherent substructures, and 
$v_\mathrm{global}$ is a virtual global node representing the entire 
molecule. $\mathcal{E}_{\mathrm{bond}}$ contains chemical 
bonds, and $\mathcal{E}_{\mathrm{hier}}$ contains directed 
edges connecting atom, fragment, and global levels. 

\textbf{Definition 2 (Joint SMILES--Fingerprint Token Sequence).} 
Let $\mathcal{R}_{\mathrm{FP}} = \{R_1, \dots, R_K\}$ denote a set of fingerprint families. 
For each active entry in family $R_k$, we construct a discrete token 
$t_p = \texttt{<fp\_name\_index\_value>}$ encoding its fingerprint type, bit index, and count value. 
The sparse fingerprint token sequence is denoted by:
\begin{equation}
    X_{\mathrm{FP}} = (t_p \mid t_p \in
    \mathrm{Active}(R_k),\ \forall R_k \in
    \mathcal{R}_{\mathrm{FP}}).
\end{equation}
Given the SMILES token sequence $X_{\mathrm{SMI}}$, the joint token sequence is defined as:
\begin{equation}
    X_{\mathrm{SF}} =
    [X_{\mathrm{SMI}} \,\|\, X_{\mathrm{FP}}],
\end{equation}
which serves as the input to the fingerprint encoder.

\textbf{Problem 1 (Multi-View Molecular Property Prediction).} 
Given a molecular dataset $\mathcal{D}=\{(M_n,y_n)\}_{n=1}^{N}$, where $y_n$ denotes the ground-truth property label, each molecule $M_n$ is represented by a hierarchical graph $\mathcal{G}_n$ and a joint token sequence $X_{\mathrm{SF},n}$. 
The goal is to learn a prediction function $f_{\mathrm{pred}}$:
\begin{equation}
    \hat{y}_n = f_{\mathrm{pred}}\bigl(
    \Phi_{\mathrm{GNN}}(\mathcal{G}_n)
    \,\|\, 
    \Phi_{\mathrm{FP}}(X_{\mathrm{SF},n})
    \bigr),
\end{equation}
where $\Phi_{\mathrm{GNN}}$ and $\Phi_{\mathrm{FP}}$ denote the graph and fingerprint encoders, respectively, $\|$ denotes feature concatenation, and $\hat{y}_n$ is the predicted molecular property.

\begin{figure*}[t]
    \centering
    \includegraphics[width=\textwidth]{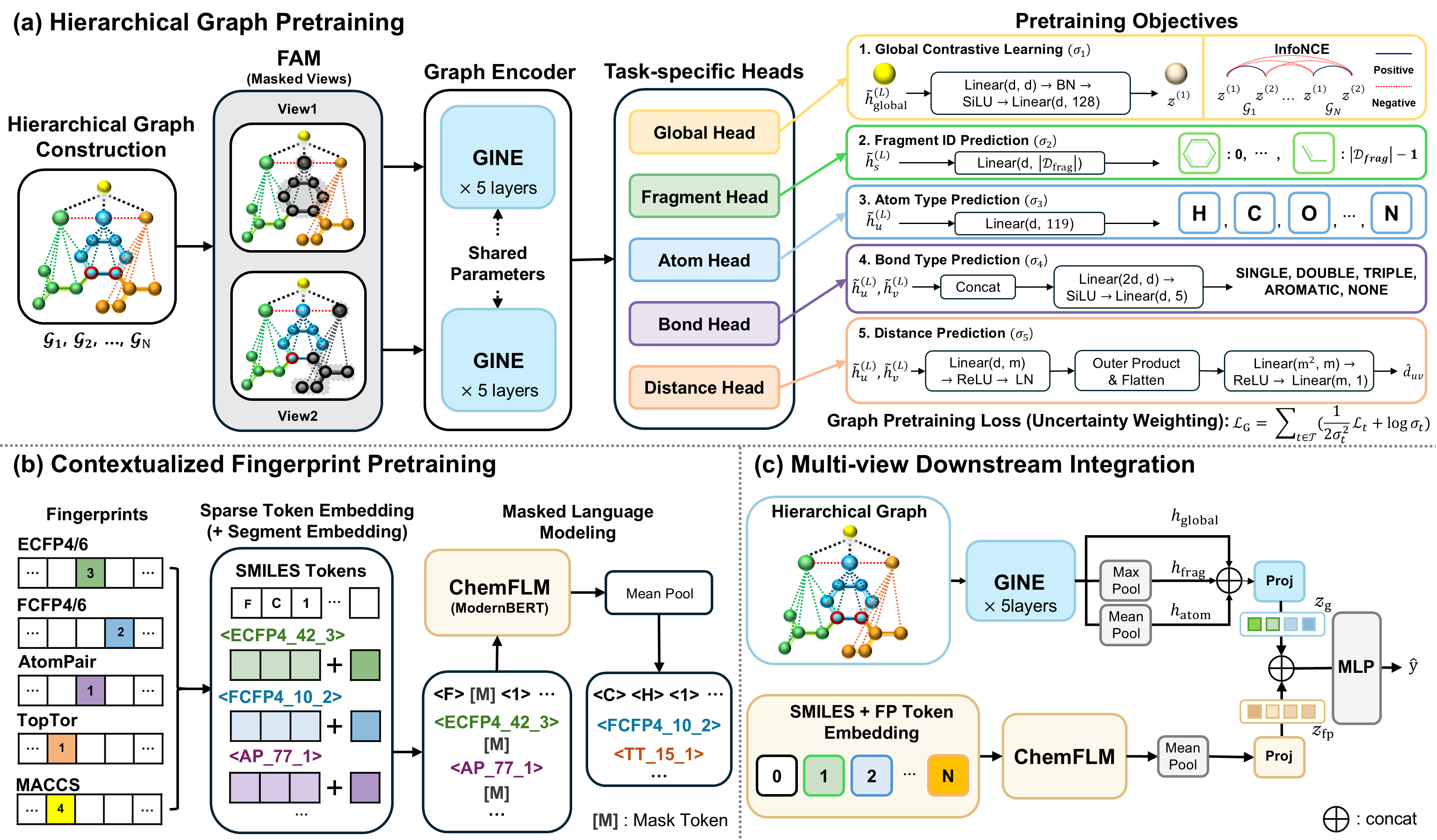}
    \caption{
        Overview of the HiFi-Mol framework. (a) Given a hierarchical molecular graph, fragment-aware masking produces two masked views that are passed through a shared 5-layer GINE encoder and optimized with global, fragment, atom, bond, and distance prediction objectives. (b) SMILES and multi-type fingerprint tokens are concatenated and processed by ChemFLM via masked language modeling. (c) For downstream prediction, the pretrained encoders provide multi-resolution graph features and contextualized fingerprint embeddings, which are projected, concatenated, and passed to a task-specific prediction head.
    }
    \label{fig:framework}
\end{figure*}

\section{Proposed Method}
Figure~\ref{fig:framework} illustrates HiFi-Mol, a multi-view pretraining framework for molecular property prediction, consisting of three stages: hierarchical graph pretraining, contextualized fingerprint pretraining, and multi-view downstream integration.

\subsection{Hierarchical Graph Pretraining}
\textbf{Hierarchical Graph Construction.} As shown in Figure~\ref{fig:fam}(a), fragment nodes $\mathcal{V}_{\mathrm{frag}}$ are constructed via a two-stage decomposition. 
The BRICS algorithm~\citep{degen2008art} first cleaves retrosynthetically meaningful bonds to isolate initial fragments. 
SSSR-based ring refinement then further partitions coarse ring-containing fragments into the smallest chemically coherent cyclic substructures. 
Each resulting fragment is canonicalized into a SMILES string and mapped to a unique ID through a dictionary function $\phi : \mathcal{S}_{\mathrm{frag}} \rightarrow \{0, \dots, |\mathcal{D}_{\mathrm{frag}}|-1\}$, where fragments occurring fewer than $\tau_{\min}=10$ times are assigned to an $\langle\mathrm{UNK}\rangle$ token.

\textbf{Fragment-aware Masking.} As illustrated in Figure~\ref{fig:fam}(b), fragment-aware masking (FAM) is applied to the hierarchical graph $\mathcal{G}$ as a corruption procedure. 
Unlike atom-level masking, which can often be inferred from local neighborhoods, FAM selects fragment nodes as masking units, requiring the encoder to recover coherent substructures rather than isolated graph elements. 
Formally, a subset of fragment nodes $\mathcal{M}_{\mathrm{frag}} \subset \mathcal{V}_{\mathrm{frag}}$ is sampled at ratio $r_f = 0.15$ and designated as masked fragments. The corresponding atom mask is defined as
\begin{equation}
    \mathcal{M}_{\mathrm{atom}} = \left\{
    v_u \in \mathcal{V}_{\mathrm{atom}} \mid
    \exists\, v_s \in \mathcal{M}_{\mathrm{frag}}
    \;\text{s.t.}\; v_u \in \mathcal{A}(s)
    \right\},
\end{equation}
where $v_u$ and $v_s$ denote the atom and fragment nodes indexed by $u$ and $s$, respectively, and $\mathcal{A}(s)$ is the set of atom nodes belonging to fragment $s$.
%where $v_s$ denotes the fragment node corresponding to fragment $s$, and $\mathcal{A}(s)$ is the set of atom nodes belonging to fragment $s$.
If the induced atom masking ratio falls below $r_a = 0.25$, additional atoms are uniformly sampled to ensure sufficient corruption. The set of masked bonds is defined as
\begin{equation}
    \mathcal{M}_{\mathrm{bond}} = \{(v_u,v_v) \in
    \mathcal{E}_{\mathrm{bond}} \mid
    v_u \in \mathcal{M}_{\mathrm{atom}}
    \text{ or } v_v \in \mathcal{M}_{\mathrm{atom}}\},
\end{equation}
where $(v_u,v_v)$ denotes a chemical bond incident to at least one masked atom. 
This produces a masked graph $\tilde{\mathcal{G}}$, in which masked nodes $\mathcal{M}_{\mathrm{atom}} \cup \mathcal{M}_{\mathrm{frag}}$ and masked bonds $\mathcal{M}_{\mathrm{bond}}$ serve as reconstruction targets for the encoder.

\textbf{Hierarchical Graph Encoding.} 
Given the masked hierarchical graph $\tilde{\mathcal{G}}$, node and edge features are represented as $h_i^{(0)} \in \mathbb{R}^d$ and $e_{ij} \in \mathbb{R}^d$, respectively. 
Node features are derived from atom attributes, fragment dictionary identities, or global molecular context depending on the node type, with masked atom and fragment nodes replaced by a learnable mask embedding. 
Edge features incorporate bond attributes and hierarchical relation types, enabling information propagation over both chemical bonds and cross-resolution connections. 
Detailed feature initialization procedures are provided in Appendix~\ref{appendix:graph_init}.

A shared GINE backbone~\citep{hu2019strategies} encodes $\tilde{\mathcal{G}}$ by performing message passing over both chemical and hierarchical edges. At each layer $l$, the representation of node $v_i$ is updated as:
\begin{equation}
    h_i^{(l)} = \mathrm{MLP}^{(l)}\!\left(
    \left(1 + \epsilon^{(l)}\right) h_i^{(l-1)} +
    \sum_{j \in \mathcal{N}(i)} \mathrm{ReLU}\!\left(
    h_j^{(l-1)} + e_{ij} \right)
    \right),
\end{equation}
where $\mathcal{N}(i)$ denotes the neighbors of $v_i$ in $\tilde{\mathcal{G}}$, and $\epsilon^{(l)}$ is a learnable scalar that controls the contribution of the previous-layer representation of $v_i$.
After $L$ layers, the encoder produces final-layer node representations $\tilde{h}_i^{(L)} \in \mathbb{R}^d$, which capture multi-resolution chemical context across atom, fragment, and global levels.

\textbf{Multi-resolution Pretraining Objectives.} The final-layer representations $\{\tilde{h}_i^{(L)}\}$ are optimized with five objectives that supervise molecule-level consistency, masked substructure recovery, local atom-bond reconstruction, and geometric regularization.

For molecule-level supervision, we generate two independently masked views, $\tilde{\mathcal{G}}^{(1)}$ and $\tilde{\mathcal{G}}^{(2)}$, for each molecule using FAM. 
Their global node embeddings are projected and $\ell_2$-normalized to obtain $z^{(1)}$ and $z^{(2)}$, which are then optimized with an InfoNCE loss~\citep{chen2020simple}:
\begin{equation}
    \mathcal{L}_{\mathrm{CL}} = -\mathbb{E}\!\left[\log
    \frac{e^{\langle z^{(1)}, z^{(2)} \rangle / \tau}}{
    e^{\langle z^{(1)}, z^{(2)} \rangle / \tau} +
    \sum_{z^- \in \mathcal{Z}^-}
    e^{\langle z^{(1)}, z^- \rangle / \tau}}
    \right],
\end{equation}
where $\tau=0.1$ is the temperature parameter that scales the similarity logits, and $\mathcal{Z}^-$ consists of in-batch negatives from different molecules.

For substructure-level supervision, masked fragment identities and atom types are recovered via linear classification heads over their respective masked node representations:
\begin{equation}
    \mathcal{L}_{\mathrm{frag}} =
    -\!\!\sum_{v_s \in \mathcal{M}_{\mathrm{frag}}}
    \log P\!\left(\phi(s) \mid \tilde{h}_s^{(L)}\right),
\end{equation}
\begin{equation}
    \mathcal{L}_{\mathrm{atom}} =
    -\!\!\sum_{v_u \in \mathcal{M}_{\mathrm{atom}}}
    \log P\!\left(a_u \mid \tilde{h}_u^{(L)}\right),
\end{equation}
while bond types are predicted by a two-layer MLP over concatenated endpoint representations:
\begin{equation}
    \mathcal{L}_{\mathrm{bond}} =
    -\!\!\sum_{(v_u,v_v) \in \mathcal{M}_{\mathrm{bond}}}
    \log P\!\left(c_{uv} \mid \tilde{h}_u^{(L)},
    \tilde{h}_v^{(L)}\right).
\end{equation}
In these losses, $\phi(s)$, $a_u$, and $c_{uv}$ denote the ground-truth fragment dictionary ID, atom type of atom node $v_u$, and bond type between atom nodes $v_u$ and $v_v$, respectively.

For geometry-level supervision, a distance decoder $f_{\mathrm{dist}}$ predicts the Euclidean distance between sampled atom pairs from their final representations:
\begin{equation}
    \mathcal{L}_{\mathrm{dist}} =
    \sum_{(v_u, v_v) \in \mathcal{P}}
    \left(f_{\mathrm{dist}}(\tilde{h}_u^{(L)}, \tilde{h}_v^{(L)}) - \|p_u - p_v\|_2\right)^2,
\end{equation}
where $\mathcal{P}$ denotes sampled atom pairs including bonded and randomly sampled non-bonded pairs, and $p_u, p_v \in \mathbb{R}^3$ are the 3D coordinates available only during pretraining.

\textbf{Uncertainty-based Loss Weighting.} To balance the five pretraining objectives without manual tuning, we adopt uncertainty-based loss weighting~\citep{kendall2018multi}. 
Let $\sigma_t^2$ denote a learnable variance parameter for objective $t$. 
The total graph pretraining loss is:
\begin{equation}
    \mathcal{L}_{\mathrm{G}}
    =
    \sum_{t \in \mathcal{T}}
    \left(
    \frac{1}{2\sigma_t^2}\mathcal{L}_t
    +
    \log \sigma_t
    \right),
\end{equation}
where $\mathcal{T}$ denotes the set of pretraining objectives. 
The first term adaptively down-weights objectives with higher uncertainty, while the log-variance term prevents trivial growth of $\sigma_t$.

\subsection{Contextualized Fingerprint Pretraining}
To complement the graph-based structural view, we independently pretrain a fingerprint encoder $\Phi_{\mathrm{FP}}$ by reframing molecular fingerprints as discrete token sequences. 
The fingerprint view is constructed from seven complementary fingerprint families capturing chemical patterns across multiple structural scales, including local circular environments (ECFP4, ECFP6), pharmacophoric features (FCFP4, FCFP6), long-range topological relations (ATOMPAIR), torsional fragments (TOPTOR), and interpretable structural keys (MACCS). 
$\Phi_{\mathrm{FP}}$ is initialized from ChemFLM, a ModernBERT-based fingerprint language model~\citep{warner2025smarter, seyed2026delbert}, enabling contextualized descriptor modeling beyond fixed fingerprint vectors.

Given the joint token sequence $X_{\mathrm{SF}}$ defined in Section~\ref{sec:preliminaries}, token and segment embeddings are summed to form the input representations:
\begin{equation}
    H_{\mathrm{SF}}^{(0)} =
    E_{\mathrm{tok}}(X_{\mathrm{SF}}) +
    E_{\mathrm{seg}}(X_{\mathrm{SF}}),
\end{equation}
where $E_{\mathrm{tok}}$ maps discrete tokens to embeddings and $E_{\mathrm{seg}}$ distinguishes whether each token belongs to the SMILES segment or one of the fingerprint descriptor families. 
The initialized sequence is processed by stacked self-attention layers:
\begin{equation}
    H_{\mathrm{SF}}^{(l)} =
    \Phi_{\mathrm{FP}}^{(l)}\!\left(
    H_{\mathrm{SF}}^{(l-1)}\right),
    \quad l = 1, \ldots, L_{\mathrm{FP}},
\end{equation}
yielding final hidden states $H_{\mathrm{SF}} = H_{\mathrm{SF}}^{(L_{\mathrm{FP}})} \in \mathbb{R}^{|X_{\mathrm{SF}}| \times d_{\mathrm{FP}}}$.

During pretraining, a masked sequence $\tilde{X}_{\mathrm{SF}}$ is constructed by randomly replacing 15\% of joint tokens with a mask token. 
Let $\tilde{H}_{\mathrm{SF}}=\Phi_{\mathrm{FP}}(\tilde{X}_{\mathrm{SF}})$ denote the final hidden states obtained from the masked input. 
For each masked position $p \in \mathcal{M}_{\mathrm{SF}}$, an MLM head predicts the original token $x_p$ from the corresponding contextualized representation $\tilde{H}_{\mathrm{SF},p}$. 
The encoder is trained to reconstruct the original tokens at masked positions:
\begin{equation}
    \mathcal{L}_{\mathrm{FP}} =
    -\sum_{p \in \mathcal{M}_{\mathrm{SF}}}
    \log P\!\left(x_p \mid \tilde{X}_{\mathrm{SF}}\right),
\end{equation}
where $x_p$ denotes the original token at masked position $p$. 
By maintaining $\Phi_{\mathrm{FP}}$ as a standalone module separate from $\Phi_{\mathrm{GNN}}$, HiFi-Mol preserves descriptor-level chemical semantics distinct from topology-driven graph representations.

\subsection{Multi-View Downstream Integration}
Since the graph and fingerprint encoders capture distinct chemical views, HiFi-Mol combines their representations only during downstream adaptation after view-specific pretraining. 
This design preserves the hierarchical structural information encoded across atom, fragment, and global levels, as well as the descriptor-level chemical knowledge captured by the fingerprint encoder.

Molecule-level representations are derived independently from each view. 
Given the hierarchical graph $\mathcal{G}$, the graph encoder $\Phi_{\mathrm{GNN}}$ yields final-layer representations at three levels: atom-level $H_{\mathrm{atom}}$, fragment-level $H_{\mathrm{frag}}$, and global-level $h_{\mathrm{global}}$. 
A molecule-level atom summary $h_{\mathrm{atom}}$ is obtained by mean pooling over $H_{\mathrm{atom}}$, a fragment summary $h_{\mathrm{frag}}$ by max pooling over $H_{\mathrm{frag}}$, and $h_{\mathrm{global}}$ is retained directly. 
For the fingerprint view, $\Phi_{\mathrm{FP}}$ processes $X_{\mathrm{SF}}$ without masking to produce final hidden states $H_{\mathrm{SF}}$. 
The molecule-level representation $h_{\mathrm{FP}}$ is obtained by mean pooling over $H_{\mathrm{SF}}$.

These view-specific summaries are then projected into compatible embedding spaces, concatenated, and passed to a task-specific prediction head: 
\begin{align}
    z_g &= f_{\mathrm{GNN}}\!\left(\left[
    h_{\mathrm{atom}} \,\|\, h_{\mathrm{frag}} \,\|\,
    h_{\mathrm{global}}\right]\right), \\
    z_{\mathrm{fp}} &= f_{\mathrm{FP}}\!\left(
    h_{\mathrm{FP}}\right), \\
    \hat{y} &= f_{\mathrm{pred}}\!\left(\left[
    z_g \,\|\, z_{\mathrm{fp}}\right]\right).
\end{align}
In this implementation, $f_{\mathrm{GNN}}$ and $f_{\mathrm{FP}}$ are two-layer MLP projection heads with batch normalization, SiLU activation, and dropout, producing 128-dimensional graph and fingerprint embeddings.  
Similarly, $f_{\mathrm{pred}}$ uses the same MLP block to map the concatenated representation to the task output.

\section{Experiments}
\subsection{Experimental Settings}

\begin{table}[t]
    \caption{Dataset statistics and downstream hyperparameter settings for MoleculeNet benchmarks. All datasets use batch size 32, weight decay 0, and dropout 0.1.}
    \label{tab:hyperparams}
    \centering
    \begin{tabular}{lcccc}
        \toprule
        \textbf{Dataset} & \textbf{Avg. \# Atoms} & \textbf{Avg. \# Frags} & \textbf{LR} & \textbf{Epochs} \\
        \midrule
        \multicolumn{5}{l}{\textit{Classification benchmarks}} \\
        \midrule
        BBBP          & 24.1 & 6.5 & $10^{-4}$ & 20  \\
        Tox21         & 18.6 & 4.9 & $10^{-5}$ & 100 \\
        ToxCast       & 18.8 & 5.1 & $10^{-4}$ & 100 \\
        HIV           & 25.5 & 6.7 & $10^{-5}$ & 20  \\
        MUV           & 24.2 & 7.1 & $10^{-4}$ & 10  \\
        ClinTox       & 26.2 & 7.0 & $10^{-5}$ & 100 \\
        BACE          & 34.1 & 9.4 & $10^{-5}$ & 20  \\
        SIDER         & 33.6 & 9.0 & $10^{-4}$ & 20  \\
        \midrule
        \multicolumn{5}{l}{\textit{Regression benchmarks}} \\
        \midrule
        ESOL          & 13.3 & 3.4 & $10^{-4}$ & 150 \\
        FreeSolv      & 8.7  & 2.2 & $10^{-4}$ & 150 \\
        Lipophilicity & 27.0 & 7.8 & $10^{-4}$ & 100 \\
        \bottomrule
    \end{tabular}
\end{table}

\begin{table*}[t]
  \caption{ROC-AUC ($\uparrow$) on MoleculeNet classification benchmarks under the scaffold split (mean $\pm$ std over 3 seeds). Best in \textbf{bold}, second-best \underline{underlined}. HiFi-Mol$_{\mathrm{Graph}}$ and HiFi-Mol$_{\mathrm{FP}}$ denote graph-only and fingerprint-only variants, respectively.}
  \label{tab:classification}
  \centering
  \begin{tabular}{lccccccccc}
    \toprule
    Dataset & BBBP & Tox21 & ToxCast & HIV & MUV & ClinTox & BACE & SIDER & Avg. AUC   \\
    \midrule
    \# Molecules & 2,039 & 7,831 & 8,577 & 41,127 & 93,087 & 1,477 & 1,513 & 1,427 & - \\ 
    \# Tasks & 1 & 12 & 617 & 1 & 17 & 2 & 1 & 27 & - \\ 
    \midrule
    AttrMask~\citep{hu2019strategies}     & 69.25\std{1.85} & 74.46\std{0.20} & 62.90\std{0.47} & 75.49\std{0.86} & 73.03\std{0.76} & 71.28\std{8.24} & 78.36\std{0.87} & 58.09\std{0.53} & 70.36 \\
    ContextPred~\citep{hu2019strategies} & 67.25\std{1.86} & 73.73\std{0.44} & 61.61\std{0.75} & 74.26\std{1.56} & 73.53\std{2.42} & 76.03\std{3.67} & 75.94\std{0.96} & 59.20\std{0.44} & 70.19 \\
    GPT-GNN~\citep{hu2020gpt} & 61.33\std{1.59} & 74.36\std{0.88} & 62.93\std{0.61} & 76.87\std{0.81} & 75.53\std{0.72} & 54.47\std{1.51} & 74.51\std{1.17} & 57.13\std{1.74} & 67.14 \\
    InfoGraph~\citep{sun2019infograph} & 68.54\std{1.39} & 73.60\std{0.79} & 60.98\std{0.95} & 75.61\std{0.59} & 74.47\std{2.94} & 69.02\std{4.65} & 61.30\std{3.41} & 58.20\std{1.10} & 67.72 \\
    GraphCL~\citep{you2020graph} & 69.65\std{1.06} & 75.17\std{0.23} & 62.83\std{0.87} & 75.72\std{1.26} & 74.30\std{1.66} & 71.56\std{8.08} & 73.00\std{0.79} & 61.14\std{0.70} & 70.42 \\
    JOAOv2~\citep{you2021graph} & 69.83\std{0.79} & 74.76\std{0.67} & 63.30\std{0.90} & 74.27\std{2.31} & 74.94\std{1.15} & 66.44\std{1.25} & 75.77\std{0.35} & 60.64\std{1.01} & 69.99 \\
    MolCLR~\citep{wang2022molecular} & \underline{72.16\std{0.34}} & 73.63\std{0.28} & 62.18\std{0.56} & \underline{77.64\std{0.95}} & 76.55\std{1.06} & 90.64\std{1.61} & \underline{82.13\std{1.06}} & 57.83\std{0.71} & 74.10 \\
    GraphMVP~\citep{liu2021pre} & 68.44\std{0.57} & 74.26\std{1.00} & 64.25\std{0.46} & 75.74\std{1.91} & 75.61\std{1.67} & 72.84\std{1.89} & 80.84\std{2.71} & 61.38\std{1.09} & 71.67 \\
    MoleculeSDE~\citep{liu2023group} & 70.68\std{0.97} & 75.04\std{0.24} & 62.62\std{0.22} & 76.89\std{0.50} & 76.22\std{1.36} & 72.45\std{3.07} & 80.74\std{0.59} & 59.82\std{0.35} & 71.81 \\
    Mole-BERT~\citep{xia2023mole} & 67.57\std{1.08} & 75.45\std{0.51} & 62.93\std{0.83} & 77.56\std{0.63} & 76.69\std{0.88} & 70.87\std{5.91} & 78.57\std{0.53} & 58.21\std{0.38} & 70.98 \\
    MGSSL~\citep{zhang2021motif} & 68.06\std{1.53} & \underline{75.65\std{0.26}} & 62.76\std{0.26} & 73.78\std{1.34} & 74.45\std{1.42} & 66.96\std{4.80} & \underline{82.13\std{1.64}} & 58.04\std{1.26} & 70.23 \\
    HiMol~\citep{zang2023hierarchical} & 70.62\std{0.61} & \textbf{76.82\std{0.83}} & \underline{65.37\std{0.35}} & 73.04\std{1.82} & \underline{76.80\std{0.97}} & 60.79\std{1.30} & 80.31\std{0.68} & 58.66\std{3.29} & 70.30 \\
    \midrule
    HiFi-Mol$_{\mathrm{Graph}}$ & 70.68\std{1.53} & 74.76\std{0.23} & \textbf{66.05\std{0.27}}
            & 75.49\std{0.81} & 74.82\std{0.61} & 88.84\std{1.91}
            & \textbf{82.16\std{0.61}} & 61.03\std{0.40} & 74.23 \\
    HiFi-Mol$_{\mathrm{FP}}$& 70.09\std{0.93} & 74.03\std{0.39} & 64.86\std{0.82}
            & \textbf{78.20\std{0.60}} & 75.62\std{1.40} & \underline{91.68\std{6.38}}
            & 80.49\std{0.90} & \textbf{64.23\std{1.39}} & \underline{74.90} \\
    HiFi-Mol & \textbf{73.68\std{0.55}} & 74.56\std{1.07} & 65.27\std{0.29} & 77.10\std{0.45} & \textbf{77.29\std{2.36}} & \textbf{96.83\std{0.95}} & 81.84\std{1.36} & \underline{62.64\std{0.70}} & \textbf{76.15} \\
    \bottomrule
  \end{tabular}%
\end{table*}

\noindent\textbf{Datasets and Metrics.}
The graph encoder is pretrained on PCQM4M-v2~\citep{hu2021ogb}, containing 3.3 million molecules with 3D coordinates used only for distance prediction. 
The fingerprint encoder is initialized from ChemFLM, pretrained on 4.4 million molecules from AIRCHECK~\citep{edwards2025protein}, ChEMBL~\citep{gaulton2012chembl}, and MOSES~\citep{polykovskiy2020molecular}. 
We evaluate on 11 MoleculeNet benchmarks under the scaffold split~\citep{wu2018moleculenet}: eight classification datasets and three regression datasets, whose statistics are summarized in Table~\ref{tab:hyperparams}.
Performance is measured by ROC-AUC for classification and RMSE for regression.

\noindent\textbf{Baselines.}
We compare against representative molecular pretraining methods from four categories: generative SSL methods (AttrMask~\citep{hu2019strategies}, ContextPred~\citep{hu2019strategies}, GPT-GNN~\citep{hu2020gpt}), contrastive SSL methods (InfoGraph~\citep{sun2019infograph}, GraphCL~\citep{you2020graph}, JOAOv2~\citep{you2021graph}, MolCLR~\citep{wang2022molecular}), hybrid SSL methods (GraphMVP~\citep{liu2021pre}, MoleculeSDE~\citep{liu2023group}, Mole-BERT~\citep{xia2023mole}), and fragment-based SSL methods (MGSSL~\citep{zhang2021motif}, HiMol~\citep{zang2023hierarchical}). Baseline results are reproduced using officially released implementations under the same scaffold split protocol and metric settings.

\noindent\textbf{Implementation Details.}
The graph encoder uses 5 GINE layers with hidden dimension 300, pretrained for 100 epochs with batch size 1,024, learning rate $10^{-4}$, AdamW optimizer, and a OneCycle scheduler with 10\% warmup followed by cosine annealing. Fragment masking ratio is set to $r_f = 0.15$, with additional random atom masking to maintain a minimum atom masking ratio of $r_a = 0.25$. The fingerprint encoder uses a ModernBERT large encoder with 12 layers and hidden size 1024, pretrained with learning rate $10^{-4}$, AdamW optimizer, batch size 64, and a cosine annealing schedule with 10\% warmup. Downstream hyperparameters are 
summarized in Table~\ref{tab:hyperparams}. All results are reported as mean and standard deviation over three random seeds. Graph pretraining is conducted on a single NVIDIA RTX A6000 GPU ($\sim$50 hours), and fingerprint encoder pretraining on a single NVIDIA RTX 4090 GPU ($\sim$500 hours).

\begin{table}[t]
    \centering
    \caption{RMSE ($\downarrow$) on MoleculeNet regression benchmarks under the scaffold split. Best in \textbf{bold}, second-best \underline{underlined}. HiFi-Mol$_{\mathrm{Graph}}$ and HiFi-Mol$_{\mathrm{FP}}$ denote graph-only and fingerprint-only variants, respectively.}
    \label{tab:regression}
    \begin{tabular}{lccc}
    \toprule
    Dataset & ESOL & FreeSolv & Lipophilicity \\
    \midrule
    \# Molecules & 1,128 & 642 & 4,200 \\ 
    \# Tasks & 1 & 1 & 1 \\ 
    \midrule
    
    MolCLR~\citep{wang2022molecular} & 1.368\std{0.012} & 2.940\std{0.099} & 0.732\std{0.014} \\
    GraphMVP~\citep{liu2021pre} & 1.294\std{0.097} & 2.741\std{0.266} & \underline{0.727\std{0.003}} \\
    Mole-BERT~\citep{xia2023mole} & 1.349\std{0.018} & 2.696\std{0.063} & 0.766\std{0.007} \\
    MGSSL~\citep{zhang2021motif} & 1.374\std{0.020} & 3.076\std{0.214} & 0.779\std{0.005} \\
    HiMol~\citep{zang2023hierarchical} & 0.884\std{0.018} & \textbf{2.042\std{0.060}} & 0.743\std{0.009} \\
    \midrule
    HiFi-Mol$_{\mathrm{Graph}}$ & \textbf{0.861\std{0.051}} & 2.644\std{0.196} & \textbf{0.718\std{0.006}} \\
    HiFi-Mol$_{\mathrm{FP}}$ & 0.968\std{0.019} & 2.454\std{0.049} & 0.830\std{0.022} \\
    HiFi-Mol & \underline{0.869\std{0.018}} & \underline{2.358\std{0.080}} & 0.743\std{0.001} \\
    \bottomrule
    \end{tabular}%
\end{table}

\subsection{Downstream Performance in MoleculeNet}
\noindent\textbf{Classification.}
Table~\ref{tab:classification} reports ROC-AUC results on eight MoleculeNet classification benchmarks. HiFi-Mol achieves the highest average ROC-AUC of 76.15, corresponding to a 2.77\% relative improvement over MolCLR, the strongest baseline. MolCLR remains competitive by learning globally consistent molecular representations, while GraphMVP and MoleculeSDE benefit from 2D-3D geometric alignment. Fragment-level methods such as MGSSL and HiMol show the value of higher-order substructure modeling, but their performance varies substantially across benchmarks. 

To analyze the contribution of each molecular view, we compare HiFi-Mol with its graph-only and fingerprint-only variants. Across the three HiFi-Mol variants, at least one achieves state-of-the-art performance on seven of the eight classification benchmarks. Within the HiFi-Mol family, the graph-only variant performs better on BBBP, Tox21, ToxCast, and BACE, while the fingerprint-only variant performs better on HIV, MUV, ClinTox, and SIDER. This dataset-dependent behavior suggests that graph and fingerprint representations provide complementary predictive signals, and their integration yields the best average performance.

\noindent\textbf{Regression.}
Table~\ref{tab:regression} reports RMSE results on three MoleculeNet regression benchmarks. 
HiFi-Mol remains competitive, achieving the second-best result on ESOL and FreeSolv, while HiFi-Mol$_{\mathrm{Graph}}$ performs best on ESOL and Lipophilicity. 
Notably, graph-based hierarchical models, including HiMol and HiFi-Mol$_{\mathrm{Graph}}$, achieve strong performance on ESOL and FreeSolv, suggesting that hierarchical structural representations can be beneficial for certain physicochemical regression tasks. 
The comparatively weaker gains from multi-view integration further suggest that some regression targets may already be sufficiently modeled by hierarchical graph representations.
%Table~\ref{tab:regression} reports RMSE results on three MoleculeNet regression benchmarks. 
%HiFi-Mol$_{\mathrm{Graph}}$ achieves the best performance on ESOL and Lipophilicity, while HiFi-Mol ranks second on ESOL and FreeSolv. Notably, hierarchical graph-based models including HiMol and HiFi-Mol$_{\mathrm{Graph}}$ consistently perform well on physicochemical regression tasks, suggesting that multi-resolution structural representations are particularly beneficial for capturing properties such as solubility and lipophilicity.

\begin{table*}[t]
    \caption{Ablation of graph pretraining components on MoleculeNet classification benchmarks under scaffold split (best in \textbf{bold}). The w/o FAM variant replaces the masking strategy with random masking; each other variant removes one pretraining objective. $\Delta$Avg. reports the change in average ROC-AUC relative to the full model.}
    \label{tab:ablation}
    \centering
    \resizebox{\textwidth}{!}{%
    \begin{tabular}{lcccccccccc}
        \toprule
        Method & BBBP & Tox21 & ToxCast & HIV & MUV 
        & ClinTox & BACE & SIDER & Avg. & $\Delta$Avg. \\
        \midrule
        \textbf{Full Model}
            & \textbf{73.68\std{0.55}} & 74.56\std{1.07} 
            & 65.27\std{0.29} & 77.10\std{0.45}
            & \textbf{77.29\std{2.36}} & 96.83\std{0.95}
            & 81.84\std{1.36} & 62.64\std{0.70} 
            & \textbf{76.15} & -\\
        w/o FAM
            & 71.19\std{1.88} & 75.05\std{0.55} 
            & 66.16\std{0.11} & 75.18\std{1.99}
            & 74.06\std{2.24} & 94.45\std{1.60} 
            & 82.45\std{0.37} & 63.37\std{1.70} 
            & 75.24 &-0.91\\
        w/o Fragment ID
            & 71.73\std{0.78} & 74.67\std{0.69} 
            & 65.61\std{0.37} & 74.63\std{0.80}
            & 74.32\std{1.47} & 97.46\std{0.94} 
            & 80.55\std{1.85} & \textbf{63.76\std{0.82}} 
            & 75.34 &-0.81\\
        w/o Distance
            & 71.63\std{0.67} & \textbf{75.27\std{0.81}} 
            & \textbf{66.37\std{0.26}} & 74.65\std{0.45}
            & 74.25\std{1.01} & \textbf{97.62\std{0.11}} 
            & 80.46\std{0.94} & 63.37\std{0.66} 
            & 75.45 &-0.70\\
        w/o Global Contrastive
            & 71.60\std{0.42} & 74.87\std{0.81} 
            & 65.76\std{0.33} & 76.30\std{1.31}
            & 73.81\std{1.54} & 94.18\std{2.65} 
            & 79.31\std{0.30} & 63.53\std{1.62} 
            & 74.92 &-1.23\\
        w/o Atom \& Bond
            & 71.01\std{1.99} & 75.00\std{0.53} 
            & 66.00\std{0.29} & \textbf{77.24\std{1.79}}
            & 75.56\std{1.67} & 96.64\std{1.76} 
            & \textbf{82.94\std{1.07}} & 62.48\std{1.10} 
            & 75.86 &-0.29\\
        \bottomrule
    \end{tabular}%
    }
\end{table*}

\begin{table}[t]
    \centering
    \caption{Effect of GNN backbone architectures within HiFi-Mol. Results report ROC-AUC under scaffold split.}
    \label{tab:backbone}
    \begin{tabular}{lcccc}
    \toprule
    Backbone & BBBP & BACE & HIV & MUV\\
    \midrule
    GIN & \textbf{73.68\std{0.55}} & 81.84\std{1.36} & \textbf{77.10\std{0.45}} & \textbf{77.29\std{2.36}} \\
    GCN & 73.45\std{0.91} & 80.82\std{0.22} & 76.36\std{1.28} & 75.68\std{0.92} \\
    GAT & 72.81\std{0.84} & 78.42\std{2.19} & 75.47\std{0.73} & 75.49\std{0.79} \\
    GraphSAGE & 72.03\std{0.92} & \textbf{82.73\std{0.29}} & 75.24\std{0.43} & 74.93\std{0.73} \\
    \bottomrule
    \end{tabular}
\end{table}

\subsection{Ablation Study}
\noindent\textbf{Graph Pretraining Components.} We analyze the contribution of each graph pretraining component through ablation on eight MoleculeNet classification benchmarks. As shown in Table~\ref{tab:ablation}, replacing fragment-aware masking with random masking reduces the average ROC-AUC by 0.91, and removing any individual objective consistently degrades performance, indicating that global contrastive consistency, fragment-level semantics, geometric regularization, and local atom-bond recovery provide complementary supervision signals. The largest degradation occurs upon removing the global contrastive objective ($\Delta\text{Avg.} = -1.23$), followed by fragment identity prediction ($-0.81$), distance prediction ($-0.70$), and atom-bond recovery ($-0.29$), whose modest impact suggests that local bonding patterns are partially recoverable via indirect supervision from the remaining objectives.

\noindent\textbf{Backbone Sensitivity.} To evaluate the generality of HiFi-Mol across different GNN architectures, we experiment with four backbone variants on a representative subset of benchmarks spanning small datasets (BBBP, BACE) and large datasets (HIV, MUV). 
As shown in Table~\ref{tab:backbone}, GIN achieves the strongest performance, which we attribute to its expressive power under the Weisfeiler-Leman graph isomorphism test~\citep{xu2018powerful}, a property particularly beneficial for distinguishing chemically distinct substructures during hierarchical pretraining. Importantly, all backbone variants remain competitive across both dataset scales, demonstrating that the hierarchical graph pretraining and fingerprint integration strategies are robust across GNN architectures with different aggregation characteristics. These results indicate that the proposed pretraining and integration strategies remain effective across multiple GNN backbones.

\begin{figure}[t]
  \centering
  \includegraphics[width=\columnwidth]{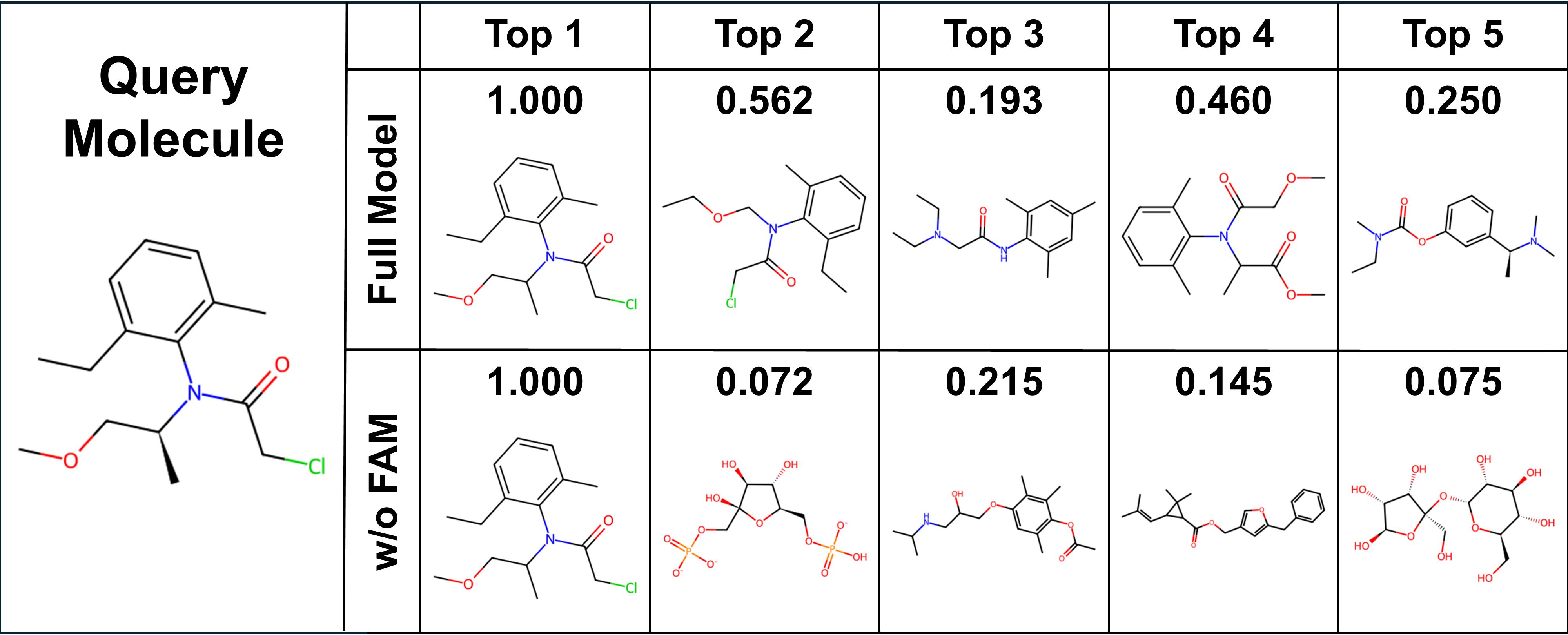}
  \caption{Top-5 retrieval results for a query molecule from the Tox21 dataset, retrieved by cosine similarity in the embedding space. Scores denote ECFP4 Tanimoto similarity to the query.}
  \label{fig:retrieval}
\end{figure}

\begin{figure}[t]
  \centering
  \includegraphics[width=\columnwidth]{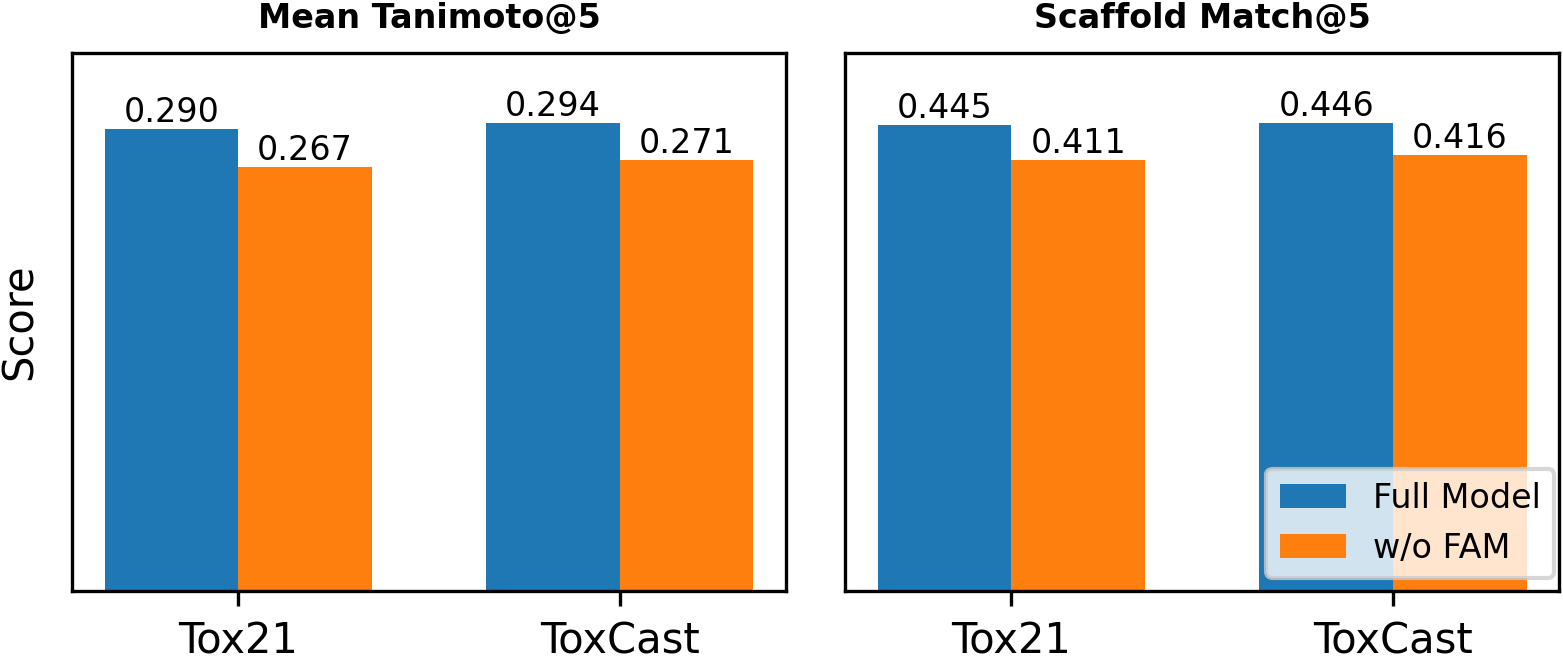}
  \caption{Quantitative comparison on Tox21 and ToxCast using Mean Tanimoto@5 and Scaffold Match@5.}
  \label{fig:quantitative}
\end{figure}
\subsection{Retrieval Analysis of Pretrained Graph Representations} 
The effect of fragment-aware masking is assessed through a molecular retrieval task on the pretrained graph encoder, prior to fingerprint integration. 
For each query molecule, we extract its embedding from the pretrained graph encoder and retrieve the top-5 nearest neighbors by cosine similarity. 
As shown in Figure~\ref{fig:retrieval}, the FAM-pretrained encoder retrieves molecules with structurally similar scaffolds to the query, with the 1st, 2nd, and 4th nearest neighbors sharing the ortho-disubstituted benzene core and maintaining considerably higher ECFP4 Tanimoto similarity. 
In contrast, the model trained without FAM retrieves chemically less coherent neighbors, including sugar derivatives and terpenoids, with substantially lower ECFP4 similarity despite high cosine similarity in the embedding space.

Retrieval quality is further quantified using Mean Tanimoto@5 and Scaffold Match@5, which measure the average ECFP4 similarity and the fraction of retrieved molecules sharing the same Bemis-Murcko scaffold~\citep{bemis1996properties} as the query, respectively. 
As illustrated in Figure~\ref{fig:quantitative}, pretraining with FAM consistently improves both metrics across Tox21 and ToxCast, with Mean Tanimoto@5 increasing by $+0.023$ on both datasets and Scaffold Match@5 improving by $+0.034$ and $+0.030$, respectively. 
Across both qualitative and quantitative evaluations, these results suggest that pretraining with FAM encourages the graph encoder to organize the embedding space along chemically meaningful structural axes, capturing scaffold-level relationships beyond local atom-level patterns.

\section{Conclusions and Future Work}
This paper presents HiFi-Mol, a multi-view molecular pretraining framework for molecular property prediction. HiFi-Mol separately pretrains hierarchical graph and contextualized fingerprint encoders before integrating them during downstream adaptation. Fragment-aware masking with multi-resolution supervision enables the graph encoder to learn substructure-aware representations, while masked language modeling over joint SMILES--fingerprint token sequences allows the fingerprint encoder to capture contextual descriptor relationships. Experiments on MoleculeNet benchmarks demonstrate the effectiveness of each design choice, with the dual-view integration achieving the best average performance across classification tasks. Nevertheless, 3D geometry is used only as pretraining supervision rather than downstream input, and the dual-view integration does not always yield gains over single-view variants on regression tasks. Future work will investigate scalable cross-view interaction mechanisms between hierarchical topology and descriptor semantics, as well as extensions to equivariant 3D molecular modeling.

\begin{acks}
This research was supported by National Research Foundation of Korea (NRF) grant funded by the Ministry of Science and ICT (RS-2025-16067916), Basic Science Research Program through the NRF funded by the Ministry of Education (RS-2025-25432868), IITP grant funded by the Ministry of Science and ICT through the Digital Columbus Project (RS-2025-02304331), and the ANCHOR program through the Gangwon ANCHOR Center funded by the Ministry of Education and the Gangwon State, Republic of Korea (2026-ANCHOR-10-006).
\end{acks}

\appendix
\section*{Appendix}
\section{Graph Initialization Details}
\label{appendix:graph_init}
Given the masked hierarchical graph $\tilde{\mathcal{G}}$, each node $v_i \in \mathcal{V}$ is initialized into a representation $h_i^{(0)} \in \mathbb{R}^d$ according to its node type. 
A learnable type embedding $E_{\mathrm{type}} \in \mathbb{R}^{3 \times d}$ distinguishes atom, fragment, and global nodes. 
Masked atom and fragment nodes, $v_i \in \mathcal{M}_{\mathrm{atom}} \cup \mathcal{M}_{\mathrm{frag}}$, are assigned a learnable mask embedding $h_{\mathrm{mask}} \in \mathbb{R}^d$ in place of their original features.

For atom nodes $v_u \in \mathcal{V}_{\mathrm{atom}}$, the initial representation is obtained by summing embedding lookups over ten chemical atom features $\{x_u^{(k)}\}_{k=1}^{10}$ (Table~\ref{tab:atom_features}):
\begin{equation}
    h_u^{(0)} = \sum_{k=1}^{10}
    E_{\mathrm{node}}^{(k)}\!\left[x_u^{(k)}\right]
    + E_{\mathrm{type}}(0).
\end{equation}
The ten atom attributes are atomic number, chirality, formal charge, hybridization, number of hydrogens, implicit valence, degree, aromaticity, ring membership, and number of radical electrons.

For fragment nodes $v_s \in \mathcal{V}_{\mathrm{frag}}$, the initial representation is determined by the fragment dictionary identity $\phi(s)$:
\begin{equation}
    h_s^{(0)} = E_{\mathrm{frag}}[\phi(s)] + E_{\mathrm{type}}(1).
\end{equation}
The global node $v_{\mathrm{global}}$ is initialized by averaging the initial atom feature embeddings:
\begin{equation}
    h_{\mathrm{global}}^{(0)} =
    \frac{1}{|\mathcal{V}_{\mathrm{atom}}|}
    \sum_{v_u \in \mathcal{V}_{\mathrm{atom}}}
    \sum_{k=1}^{10}
    E_{\mathrm{node}}^{(k)}[x_u^{(k)}]
    + E_{\mathrm{type}}(2).
\end{equation}

Each edge $(v_i,v_j)$ is initialized as the sum of four edge feature embeddings:
\begin{equation}
    e_{ij} =
    \sum_{r=1}^{4}
    E_{\mathrm{edge}}^{(r)}[b_{ij}^{(r)}],
\end{equation}
where $\{b_{ij}^{(r)}\}_{r=1}^{4}$ denotes edge attributes for bond type, bond direction, stereo configuration, and conjugation. The bond type attribute is extended beyond physical bond types to include self-loops, mask tokens, and directed hierarchical relation types, including atom-to-fragment, fragment-to-atom, fragment-to-fragment, fragment-to-global, and global-to-fragment edges.

\begin{table}[t]
    \centering
    \caption{Atom features used for node initialization. Atomic number 119 is reserved as a mask token.}
    \label{tab:atom_features}
    \begin{tabular}{@{}lc@{}}
        \toprule
        \textbf{Feature} & \textbf{Size} \\
        \midrule
        Atomic number & 119 \\
        Chirality & 4 \\
        Formal charge & 11 \\
        Hybridization & 7 \\
        Number of hydrogens & 9 \\
        Implicit valence & 7 \\
        Degree & 11 \\
        Aromaticity & 2 \\
        Ring membership & 2 \\
        Radical electrons & 5 \\
        \bottomrule
    \end{tabular}
\end{table}

\section*{GenAI Usage Disclosure}
LLM was used solely for grammar checking and language polishing of the manuscript. All scientific content, experimental design, analysis, and conclusions are entirely the work of the authors.

\bibliographystyle{ACM-Reference-Format}
\balance
\bibliography{reference}

@inproceedings{gilmer2017neural,
  title={Neural message passing for quantum chemistry},
  author={Gilmer, Justin and Schoenholz, Samuel S and Riley, Patrick F and Vinyals, Oriol and Dahl, George E},
  booktitle={International conference on machine learning},
  pages={1263--1272},
  year={2017},
  organization={PMLR}
}

@article{liu2025molecular,
  title={Molecular Motif Learning as a pretraining objective for molecular property prediction},
  author={Liu, Ziyang and Wang, Chaokun and Zheng, Shuwen and Wu, Cheng and Feng, Hao and Xu, Li and Zheng, Yue and Rong, Liang and Li, Peng},
  journal={Nature Communications},
  year={2025},
  publisher={Nature Publishing Group UK London}
}

@article{stokes2020deep,
  title={A deep learning approach to antibiotic discovery},
  author={Stokes, Jonathan M and Yang, Kevin and Swanson, Kyle and Jin, Wengong and Cubillos-Ruiz, Andres and Donghia, Nina M and MacNair, Craig R and French, Shawn and Carfrae, Lindsey A and Bloom-Ackermann, Zohar and others},
  journal={Cell},
  volume={180},
  number={4},
  pages={688--702},
  year={2020},
  publisher={Elsevier}
}

@article{hu2019strategies,
  title={Strategies for pre-training graph neural networks},
  author={Hu, Weihua and Liu, Bowen and Gomes, Joseph and Zitnik, Marinka and Liang, Percy and Pande, Vijay and Leskovec, Jure},
  journal={arXiv preprint arXiv:1905.12265},
  year={2019}
}

@inproceedings{hu2020gpt,
  title={Gpt-gnn: Generative pre-training of graph neural networks},
  author={Hu, Ziniu and Dong, Yuxiao and Wang, Kuansan and Chang, Kai-Wei and Sun, Yizhou},
  booktitle={Proceedings of the 26th ACM SIGKDD international conference on knowledge discovery \& data mining},
  pages={1857--1867},
  year={2020}
}

@article{you2020graph,
  title={Graph contrastive learning with augmentations},
  author={You, Yuning and Chen, Tianlong and Sui, Yongduo and Chen, Ting and Wang, Zhangyang and Shen, Yang},
  journal={Advances in neural information processing systems},
  volume={33},
  pages={5812--5823},
  year={2020}
}

@inproceedings{you2021graph,
  title={Graph contrastive learning automated},
  author={You, Yuning and Chen, Tianlong and Shen, Yang and Wang, Zhangyang},
  booktitle={International conference on machine learning},
  pages={12121--12132},
  year={2021},
  organization={PMLR}
}

@article{sun2019infograph,
  title={Infograph: Unsupervised and semi-supervised graph-level representation learning via mutual information maximization},
  author={Sun, Fan-Yun and Hoffmann, Jordan and Verma, Vikas and Tang, Jian},
  journal={arXiv preprint arXiv:1908.01000},
  year={2019}
}

@article{wang2022molecular,
  title={Molecular contrastive learning of representations via graph neural networks},
  author={Wang, Yuyang and Wang, Jianren and Cao, Zhonglin and Barati Farimani, Amir},
  journal={Nature Machine Intelligence},
  volume={4},
  number={3},
  pages={279--287},
  year={2022},
  publisher={Nature Publishing Group UK London}
}

@article{liu2021pre,
  title={Pre-training molecular graph representation with 3d geometry},
  author={Liu, Shengchao and Wang, Hanchen and Liu, Weiyang and Lasenby, Joan and Guo, Hongyu and Tang, Jian},
  journal={arXiv preprint arXiv:2110.07728},
  year={2021}
}

@inproceedings{liu2023group,
  title={A group symmetric stochastic differential equation model for molecule multi-modal pretraining},
  author={Liu, Shengchao and Du, Weitao and Ma, Zhi-Ming and Guo, Hongyu and Tang, Jian},
  booktitle={International Conference on Machine Learning},
  pages={21497--21526},
  year={2023},
  organization={PMLR}
}

@article{zhang2021motif,
  title={Motif-based graph self-supervised learning for molecular property prediction},
  author={Zhang, Zaixi and Liu, Qi and Wang, Hao and Lu, Chengqiang and Lee, Chee-Kong},
  journal={Advances in Neural Information Processing Systems},
  volume={34},
  pages={15870--15882},
  year={2021}
}

@article{zang2023hierarchical,
  title={Hierarchical molecular graph self-supervised learning for property prediction},
  author={Zang, Xuan and Zhao, Xianbing and Tang, Buzhou},
  journal={Communications Chemistry},
  volume={6},
  number={1},
  pages={34},
  year={2023},
  publisher={Nature Publishing Group UK London}
}

@inproceedings{warner2025smarter,
  title={Smarter, better, faster, longer: A modern bidirectional encoder for fast, memory efficient, and long context finetuning and inference},
  author={Warner, Benjamin and Chaffin, Antoine and Clavi{\'e}, Benjamin and Weller, Orion and Hallstr{\"o}m, Oskar and Taghadouini, Said and Gallagher, Alexis and Biswas, Raja and Ladhak, Faisal and Aarsen, Tom and others},
  booktitle={Proceedings of the 63rd Annual Meeting of the Association for Computational Linguistics (Volume 1: Long Papers)},
  pages={2526--2547},
  year={2025}
}

@inproceedings{xia2023mole,
  title={Mole-bert: Rethinking pre-training graph neural networks for molecules},
  author={Xia, Jun and Zhao, Chengshuai and Hu, Bozhen and Gao, Zhangyang and Tan, Cheng and Liu, Yue and Li, Siyuan and Li, Stan Z},
  booktitle={The Eleventh International Conference on Learning Representations},
  year={2023}
}

@article{xu2018powerful,
  title={How powerful are graph neural networks?},
  author={Xu, Keyulu and Hu, Weihua and Leskovec, Jure and Jegelka, Stefanie},
  journal={arXiv preprint arXiv:1810.00826},
  year={2018}
}

@inproceedings{chen2020simple,
  title={A simple framework for contrastive learning of visual representations},
  author={Chen, Ting and Kornblith, Simon and Norouzi, Mohammad and Hinton, Geoffrey},
  booktitle={International conference on machine learning},
  pages={1597--1607},
  year={2020},
  organization={PMLR}
}

@inproceedings{kendall2018multi,
  title={Multi-task learning using uncertainty to weigh losses for scene geometry and semantics},
  author={Kendall, Alex and Gal, Yarin and Cipolla, Roberto},
  booktitle={Proceedings of the IEEE conference on computer vision and pattern recognition},
  pages={7482--7491},
  year={2018}
}

@article{hu2021ogb,
  title={Ogb-lsc: A large-scale challenge for machine learning on graphs},
  author={Hu, Weihua and Fey, Matthias and Ren, Hongyu and Nakata, Maho and Dong, Yuxiao and Leskovec, Jure},
  journal={arXiv preprint arXiv:2103.09430},
  year={2021}
}

@article{wu2018moleculenet,
  title={MoleculeNet: a benchmark for molecular machine learning},
  author={Wu, Zhenqin and Ramsundar, Bharath and Feinberg, Evan N and Gomes, Joseph and Geniesse, Caleb and Pappu, Aneesh S and Leswing, Karl and Pande, Vijay},
  journal={Chemical science},
  volume={9},
  number={2},
  pages={513--530},
  year={2018},
  publisher={Royal Society of Chemistry}
}

@article{rong2020self,
  title={Self-supervised graph transformer on large-scale molecular data},
  author={Rong, Yu and Bian, Yatao and Xu, Tingyang and Xie, Weiyang and Wei, Ying and Huang, Wenbing and Huang, Junzhou},
  journal={Advances in neural information processing systems},
  volume={33},
  pages={12559--12571},
  year={2020}
}

@inproceedings{stark20223d,
  title={3d infomax improves gnns for molecular property prediction},
  author={St{\"a}rk, Hannes and Beaini, Dominique and Corso, Gabriele and Tossou, Prudencio and Dallago, Christian and G{\"u}nnemann, Stephan and Li{\`o}, Pietro},
  booktitle={International conference on machine learning},
  pages={20479--20502},
  year={2022},
  organization={PMLR}
}

@article{jiang2023pharmacophoric,
  title={Pharmacophoric-constrained heterogeneous graph transformer model for molecular property prediction},
  author={Jiang, Yinghui and Jin, Shuting and Jin, Xurui and Xiao, Xianglu and Wu, Wenfan and Liu, Xiangrong and Zhang, Qiang and Zeng, Xiangxiang and Yang, Guang and Niu, Zhangming},
  journal={Communications Chemistry},
  volume={6},
  number={1},
  pages={60},
  year={2023},
  publisher={Nature Publishing Group UK London}
}

@article{yang2019analyzing,
  title={Analyzing learned molecular representations for property prediction},
  author={Yang, Kevin and Swanson, Kyle and Jin, Wengong and Coley, Connor and Eiden, Philipp and Gao, Hua and Guzman-Perez, Angel and Hopper, Timothy and Kelley, Brian and Mathea, Miriam and others},
  journal={Journal of chemical information and modeling},
  volume={59},
  number={8},
  pages={3370--3388},
  year={2019},
  publisher={ACS Publications}
}

@article{luong2023fragment,
  title={Fragment-based pretraining and finetuning on molecular graphs},
  author={Luong, Kha-Dinh and Singh, Ambuj K},
  journal={Advances in Neural Information Processing Systems},
  volume={36},
  pages={17584--17601},
  year={2023}
}

@article{cai2022fp,
  title={FP-GNN: a versatile deep learning architecture for enhanced molecular property prediction},
  author={Cai, Hanxuan and Zhang, Huimin and Zhao, Duancheng and Wu, Jingxing and Wang, Ling},
  journal={Briefings in bioinformatics},
  volume={23},
  number={6},
  pages={bbac408},
  year={2022},
  publisher={Oxford University Press}
}

@article{jiang2024dgcl,
  title={DGCL: dual-graph neural networks contrastive learning for molecular property prediction},
  author={Jiang, Xiuyu and Tan, Liqin and Zou, Qingsong},
  journal={Briefings in Bioinformatics},
  volume={25},
  number={6},
  pages={bbae474},
  year={2024},
  publisher={Oxford University Press}
}

@article{degen2008art,
  title={On the art of compiling and using'drug-like'chemical fragment spaces},
  author={Degen, Jorg and Wegscheid-Gerlach, Christof and Zaliani, Andrea and Rarey, Matthias},
  journal={ChemMedChem},
  volume={3},
  number={10},
  pages={1503},
  year={2008}
}

@inproceedings{seyed2026delbert,
  title={DELBERT: Fingerprint Language Modeling For Generalizable Hit Discovery in DNA-Encoded Libraries},
  author={Seyed-Ahmadi, Arman and Hu, Bing Xu and Geraili, Armin and Layton, Anita and Chen, Helen Hong and Kelley, Shana O and Wang, Bo},
  booktitle={ICLR 2026 Workshop on Machine Learning for Genomics Explorations},
  year={2026}
}

@article{gaulton2012chembl,
  title={ChEMBL: a large-scale bioactivity database for drug discovery},
  author={Gaulton, Anna and Bellis, Louisa J and Bento, A Patricia and Chambers, Jon and Davies, Mark and Hersey, Anne and Light, Yvonne and McGlinchey, Shaun and Michalovich, David and Al-Lazikani, Bissan and others},
  journal={Nucleic acids research},
  volume={40},
  number={D1},
  pages={D1100--D1107},
  year={2012},
  publisher={Oxford University Press}
}

@article{polykovskiy2020molecular,
  title={Molecular sets (MOSES): a benchmarking platform for molecular generation models},
  author={Polykovskiy, Daniil and Zhebrak, Alexander and Sanchez-Lengeling, Benjamin and Golovanov, Sergey and Tatanov, Oktai and Belyaev, Stanislav and Kurbanov, Rauf and Artamonov, Aleksey and Aladinskiy, Vladimir and Veselov, Mark and others},
  journal={Frontiers in pharmacology},
  volume={11},
  pages={565644},
  year={2020},
  publisher={Frontiers}
}

@article{edwards2025protein,
  title={Protein--ligand data at scale to support machine learning},
  author={Edwards, Aled M and Owen, Dafydd R},
  journal={Nature Reviews Chemistry},
  volume={9},
  number={9},
  pages={634--645},
  year={2025},
  publisher={Nature Publishing Group UK London}
}

@article{rogers2010extended,
  title={Extended-connectivity fingerprints},
  author={Rogers, David and Hahn, Mathew},
  journal={Journal of chemical information and modeling},
  volume={50},
  number={5},
  pages={742--754},
  year={2010},
  publisher={ACS Publications}
}

@article{deng2023systematic,
  title={A systematic study of key elements underlying molecular property prediction},
  author={Deng, Jianyuan and Yang, Zhibo and Wang, Hehe and Ojima, Iwao and Samaras, Dimitris and Wang, Fusheng},
  journal={Nature Communications},
  volume={14},
  number={1},
  pages={6395},
  year={2023},
  publisher={Nature Publishing Group UK London}
}

@article{atz2021geometric,
  title={Geometric deep learning on molecular representations},
  author={Atz, Kenneth and Grisoni, Francesca and Schneider, Gisbert},
  journal={Nature Machine Intelligence},
  volume={3},
  number={12},
  pages={1023--1032},
  year={2021},
  publisher={Nature Publishing Group UK London}
}

@article{david2020molecular,
  title={Molecular representations in AI-driven drug discovery: a review and practical guide},
  author={David, Laurianne and Thakkar, Amol and Mercado, Roc{\'\i}o and Engkvist, Ola},
  journal={Journal of cheminformatics},
  volume={12},
  number={1},
  pages={56},
  year={2020},
  publisher={Springer}
}

@article{bemis1996properties,
  title={The properties of known drugs. 1. Molecular frameworks},
  author={Bemis, Guy W and Murcko, Mark A},
  journal={Journal of medicinal chemistry},
  volume={39},
  number={15},
  pages={2887--2893},
  year={1996},
  publisher={ACS Publications}
}

@inproceedings{devlin2019bert,
  title={Bert: Pre-training of deep bidirectional transformers for language understanding},
  author={Devlin, Jacob and Chang, Ming-Wei and Lee, Kenton and Toutanova, Kristina},
  booktitle={Proceedings of the 2019 conference of the North American chapter of the association for computational linguistics: human language technologies, volume 1 (long and short papers)},
  pages={4171--4186},
  year={2019}
}

@article{duvenaud2015convolutional,
  title={Convolutional networks on graphs for learning molecular fingerprints},
  author={Duvenaud, David K and Maclaurin, Dougal and Iparraguirre, Jorge and Bombarell, Rafael and Hirzel, Timothy and Aspuru-Guzik, Al{\'a}n and Adams, Ryan P},
  journal={Advances in neural information processing systems},
  volume={28},
  year={2015}
}

@article{qiao2025self,
  title={A self-conformation-aware pre-training framework for molecular property prediction with substructure interpretability},
  author={Qiao, Jianbo and Jin, Junru and Wang, Ding and Teng, Saisai and Zhang, Junyu and Yang, Xuetong and Liu, Yuhang and Wang, Yu and Cui, Lizhen and Zou, Quan and others},
  journal={Nature Communications},
  volume={16},
  number={1},
  pages={4382},
  year={2025},
  publisher={Nature Publishing Group UK London}
}

@article{chu2026memol,
title = {MEMOL: Mixture of experts for multimodal learning through multi-head attention to predict drug toxicity},
author = {Jae-Woo Chu and Jong-Hoon Park and Young-Rae Cho},
journal = {Computer Methods and Programs in Biomedicine},
volume = {273},
pages = {109088},
year = {2026},
issn = {0169-2607},
}

@inproceedings{li2025contextual,
  title={Contextual Representation Anchor Network for Mitigating Selection Bias in Few-Shot Drug Discovery},
  author={Li, Ruifeng and Liu, Wei and Zhou, Xiangxin and Li, Mingqian and Zhang, Qiang and Chen, Hongyang and Lin, Xuemin},
  booktitle={Proceedings of the 34th ACM International Conference on Information and Knowledge Management},
  pages={1634--1642},
  year={2025}
}

@article{antoniuk2026boom,
  title={Boom: benchmarking out-of-distribution molecular property predictions of machine learning models},
  author={Antoniuk, Evan and Zaman, Shehtab and Ben-Nun, Tal and Li, Peggy and Diffenderfer, James and Sahin, Busra and Smolenski, Obadiah and Grethel, Everett and Hsu, Tim and Hiszpanski, Anna and others},
  journal={Advances in Neural Information Processing Systems},
  volume={38},
  year={2026}
}

@inproceedings{liu2025subgraph,
  title={Subgraph aggregation for out-of-distribution generalization on graphs},
  author={Liu, Bowen and Li, Haoyang and Wang, Shuning and Nie, Shuo and Zhang, Shanghang},
  booktitle={Proceedings of the AAAI Conference on Artificial Intelligence},
  volume={39},
  number={18},
  pages={18763--18771},
  year={2025}
}
\end{document}